\documentclass[conference]{IEEEtran}
\IEEEoverridecommandlockouts
\usepackage{cite}
\usepackage{amsmath,amssymb,amsfonts}
\usepackage{algorithmic}
\usepackage{graphicx}
\usepackage{textcomp}
\usepackage{xcolor}
\usepackage{booktabs}
\usepackage{tabularx} % 导入tabularx包
\usepackage{multirow}
\usepackage{tikz}
\usepackage{adjustbox}
\newcommand{\mybullet}{\tikz\draw[black,fill=black] (0,0) circle (4pt);}
\newcommand{\mycirc}{\tikz\draw[black,fill=white] (0,0) circle (4pt);}
\def\BibTeX{{\rm B\kern-.05em{\sc i\kern-.025em b}\kern-.08em
    T\kern-.1667em\lower.7ex\hbox{E}\kern-.125emX}}
\begin{document}

\title{Toward Reliable RGB-D Semantic Segmentation: Handling Missing Modalities via Condition Dropout}

\author{
\IEEEauthorblockN{
Xuchen Zhu$^{1,\dagger}$,
Yajuan Wei$^{1,\dagger}$,
Shuang Hao$^{2}$,
Jiwei Jiang$^{3}$,
Guanxiang Mao$^{4}$,
Fang Ren$^{1,*}$
}

\IEEEauthorblockA{
$^{1}$ Xi'an University of Posts \& Telecommunications, Xi'an, China \\
$^{2}$ Xi'an Jiaotong University, Xi'an, China \\
$^{3}$ Huazhong University of Science and Technology, Wuhan, China \\
$^{4}$ Beijing University of Posts and Telecommunications, Beijing, China
}

\IEEEauthorblockA{
xchzhu@stu.xupt.edu.cn,
renfang\_81@163.com
}

\thanks{$^{\dagger}$ Equal contribution.}
\thanks{$^{*}$ Corresponding author.}
}

\maketitle

%%%%%%%%%%%%%%%%%%%%%%%%%%%%%%%%%%%%%%%%%%%%%%%%%%%%%%%%%%%%%%%%%%%%%%%%%%%%%%%%%%%%%%%%%%%
\begin{abstract}
RGB-D semantic segmentation has achieved remarkable progress, yet most models assume that RGB and depth are always available. In practice, failures or occlusions of surveillance sensors often remove one modality. Although RGB or depth alone can contain sufficient cues, models trained only on full-modality inputs fail to exploit the remaining modality once one is missing, causing severe degradation. We tackle this issue with a simple continued-training paradigm, \emph{Condition Dropout (ConD)}, which mitigates degradation while preserving full-modality accuracy. Starting from a pretrained RGB-D model, ConD adds a second stage that randomly simulates complete, RGB-missing, and depth-missing inputs, freezes the original encoders, and trains copied encoders with zero-initialized feature injection. Experiments on NYU-Depth V2 and SUN RGB-D show that ConD improves robustness under missing modalities and even yields slight gains when modalities are complete. Our code will be made publicly available upon acceptance.
\end{abstract}

\begin{IEEEkeywords}
RGB-D semantic segmentation; modality missing; robustness 
%\url{http://www.ieee.org/organizations/pubs/ani_prod/keywrd98.txt}
\end{IEEEkeywords}

%%%%%%%%%%%%%%%%%%%%%%%%%%%%%%%%%%%%%%%%%%%%%%%%%%%%%%%%%%%%%%%%%%%%%%%%%%%%%%%%%%%%%%%%%%%
\section{Introduction}
\label{sec:intro}

% 我面对的主要问题是：当前已经有了大量的 rgb-d 实例分割的工作，但是对于可能因为传感器损坏等原因导致某一种模态缺失这一在现实场景中常见的情况考虑甚少。
% 我们针对可能出现的模态缺失的问题，提出了一种继续训练的策略：在已经训练好的模型基础上再进行模态缺失下的训练，可以在不损失模态完全性能的前提下，提升模态缺失时的性能。
% 实验表明我们的方法在模态缺失的情况下表现出了优异的性能，并且在模态完全的情况下也保持基本不下降的性能。
% 这项研究说明了在 rgb-d 实例分割领域引入模态缺失考量的重要性，为后续的研究提供了一个不同的视角。
RGB-D semantic segmentation plays a vital role in a wide range of real-world applications, including autonomous robotics, scene understanding, and surveillance data processing. By leveraging the complementary strengths of RGB images and depth maps, many existing works have achieved promising results through carefully designed fusion architectures. In typical approaches, RGB and depth modalities are simultaneously fed into dual-branch networks or shared encoders, and features are fused at multiple levels to improve segmentation accuracy~\cite{bai2025dcanet,erep2024esenet,du2024asymformer, zhang2021non, hu2019acnet,wang2021brief,wang2018depth,gupta2014learning}. Importantly, either RGB or depth alone often carries sufficient semantic cues for reasonable segmentation in many scenes, suggesting that a robust RGB-D model should still produce plausible predictions even when one modality is absent.

However, most existing methods are built upon the assumption that both modalities are always available and reliable, which is often not the case in real-world scenarios. In practice, due to sensor failure, occlusion, misalignment, or even being attacked, one modality (typically depth) can be missing or severely degraded. Under such modality-incomplete inputs, current RGB-D models—trained exclusively on complete-modal data—tend to output extremely poor segmentation results, far worse than what could be expected from the remaining modality alone. This indicates that their lack of robustness to missing modalities prevents them from fully exploiting the surviving modality and fundamentally limits their representational capacity. Although previous studies have focused mainly on improving fusion mechanisms under complete-modal inputs~\cite{lu2025spcformer,long2015fully, ronneberger2015u, chen2017deeplab}, the problem of missing modality remains largely underexplored~\cite{liao2025benchmarking,maheshwari2024missing,  zhao2024maskmentor,qu2024mmpl, li2023towards, zhao2023mitigating}, which in turn restricts the deployment of RGB-D models in safety-critical or resource-constrained environments where sensor reliability cannot be guaranteed.
\begin{figure}[t]
    \centering
    \includegraphics[width=\linewidth]{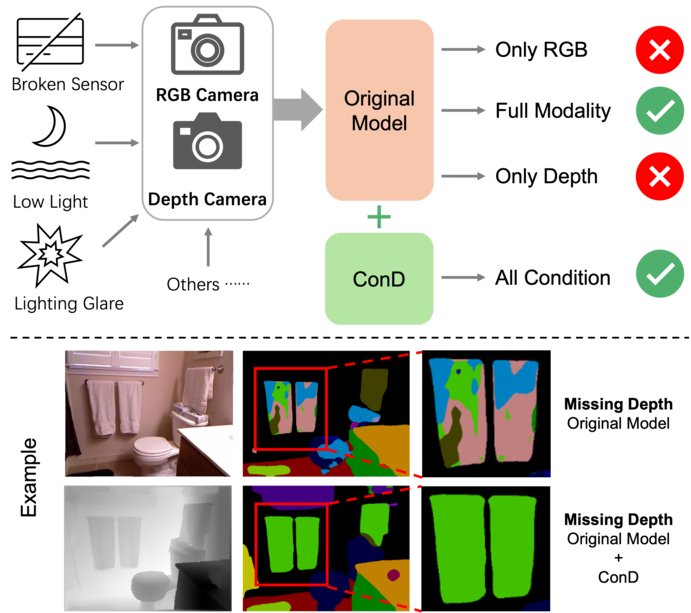}
    \caption{
            Illustration of the modality-missing problem and the effect of our method.Real-world RGB-D systems often suffer from partial modality failure (e.g., broken sensors, low light, lighting glare), which causes a severe performance drop for existing RGB-D segmentation models that are trained only with full-modality inputs. By augmenting a pretrained model with our simple Conditional Dropout (ConD) module, we can maintain stable performance across different input conditions (RGB only, depth only, and full RGB-D) without retraining the original backbone. Bottom: a qualitative example on DFormer under a missing-depth setting. Although the RGB image still contains sufficient semantic cues, the original DFormer fails to produce correct segmentation, whereas DFormer + ConD yields accurate and clean predictions.
    }
    \label{fig:task}
\end{figure}

As illustrated in Fig.~\ref{fig:task}, our goal is to achieve robust segmentation under all input conditions, including cases where one modality is missing. Inspired by recent works that explicitly consider modality missing, such as CoLA~\cite{Hao2024CoLA}, we propose a simple continued-training strategy, termed \emph{Condition Dropout (ConD)}, which augments existing RGB-D models with an additional ConD module and further trains them with incomplete-modal inputs (RGB-only or depth-only). Crucially, this module can be seamlessly attached to a wide range of pretrained RGB-D segmentation architectures without modifying their original backbones or retraining them from scratch. By explicitly exposing the model–ConD combination to modality-missing cases in this second stage, the network learns to exploit the remaining modality more effectively and drastically reduces the severe performance drop that conventional RGB-D models exhibit when one modality is absent, while preserving their accuracy under complete inputs. Extensive experiments on public RGB-D benchmarks show that our approach consistently improves robustness under missing modalities and maintains competitive, often comparable or even slightly better, accuracy under full-modal conditions, validating its effectiveness and generalizability.  

The main contributions of this work are summarized as follows:
\begin{itemize}
  \item We identify and systematically study the practical challenge of modality missing in RGB-D semantic segmentation, showing that even strong RGB-D models can produce extremely poor predictions once a modality is absent, despite the remaining modality still containing sufficient semantic cues;
  \item We propose a simple yet effective continued-training strategy with an explicit \emph{Condition Dropout (ConD)} module that can be directly plugged into existing RGB-D semantic segmentation models, adapting a fully trained backbone to handle incomplete-modal inputs (RGB-only or depth-only) without architectural changes to the original network or retraining it from scratch;
  \item We conduct comprehensive experiments on multiple standard benchmarks, demonstrating that the proposed ConD-augmented models significantly alleviate the performance degradation under missing-modality conditions while maintaining competitive performance on complete-modal inputs, thereby enabling more reliable deployment of RGB-D models in practice.
\end{itemize}

%%%%%%%%%%%%%%%%%%%%%%%%%%%%%%%%%%%%%%%%%%%%%%%%%%%%%%%%%%%%%%%%%%%%%%%%%%%%%%%%%%%%%%%%%%%
\section{Method}
% 我们的方法是作用在训练好的模型上，接着进行训练的。首先我们需要把已经训练好模型的编码器复制出来一份，并将原来的部分冻结，接下来我们在训练的时候，输入存在三种可能：只有RGB 模态，只有 Depth 模态，两个模态完全。三个组合采用等可能的随机输入方式进行输入。接下来将复制出来编码器的输出，先通过一个零卷积，再加和到原来的编码器提取出来的特征向量上面去。这里零卷积的做法我们参考了 ControlNet 的实现。
% 这么做的原理是，冻结的编码器参数保证了模型在全模态输入的效果稳定，复制出来的编码器因为先通过了零卷积，所以对模型在初始阶段没有很大的影响，保证了模型的收敛；接下来在训练的时候，缺失模态的输入使得模型更加深入的学习了每个模态更深层次的特征，从而在只输入单个模态的时候，依旧能够有较稳定的效果，同时冻结的编码器保证全模态的效果下限。
In this section, we present our proposed training paradigm designed to improve the robustness of RGB-D semantic segmentation models under modality-missing conditions. Rather than introducing architectural modifications, our approach operates at the training level and can be seamlessly applied to existing RGB-D segmentation models. \emph{Crucially, the paradigm is compute-efficient: it starts from any off-the-shelf model checkpoint and thus requires training only the second stage.}

\subsection{Overview of the Training Paradigm}

Let $\mathcal{M}_\text{pre}$ denote a pre-trained RGB-D semantic segmentation model with encoder $E_\text{pre}$ and decoder $D_\text{pre}$. \emph{Stage~1 is assumed to be already completed by existing training pipelines, providing a readily available checkpoint $\mathcal{M}_\text{pre}$.} Our paradigm initializes from $\mathcal{M}_\text{pre}$ and introduces a continued training stage aimed at enhancing the model’s capability to handle incomplete-modal inputs without compromising its full-modal performance.

To this end, we duplicate the encoder to obtain a trainable counterpart $E_\text{aux}$ initialized with the same weights:
\begin{equation}
E_\text{aux} \leftarrow \text{copy}(E_\text{pre}),\quad \theta_\text{pre} \text{ frozen},\ \theta_\text{aux} \text{ trainable}.
\end{equation}

\begin{figure}[t]
    \centering
    \includegraphics[width=\linewidth]{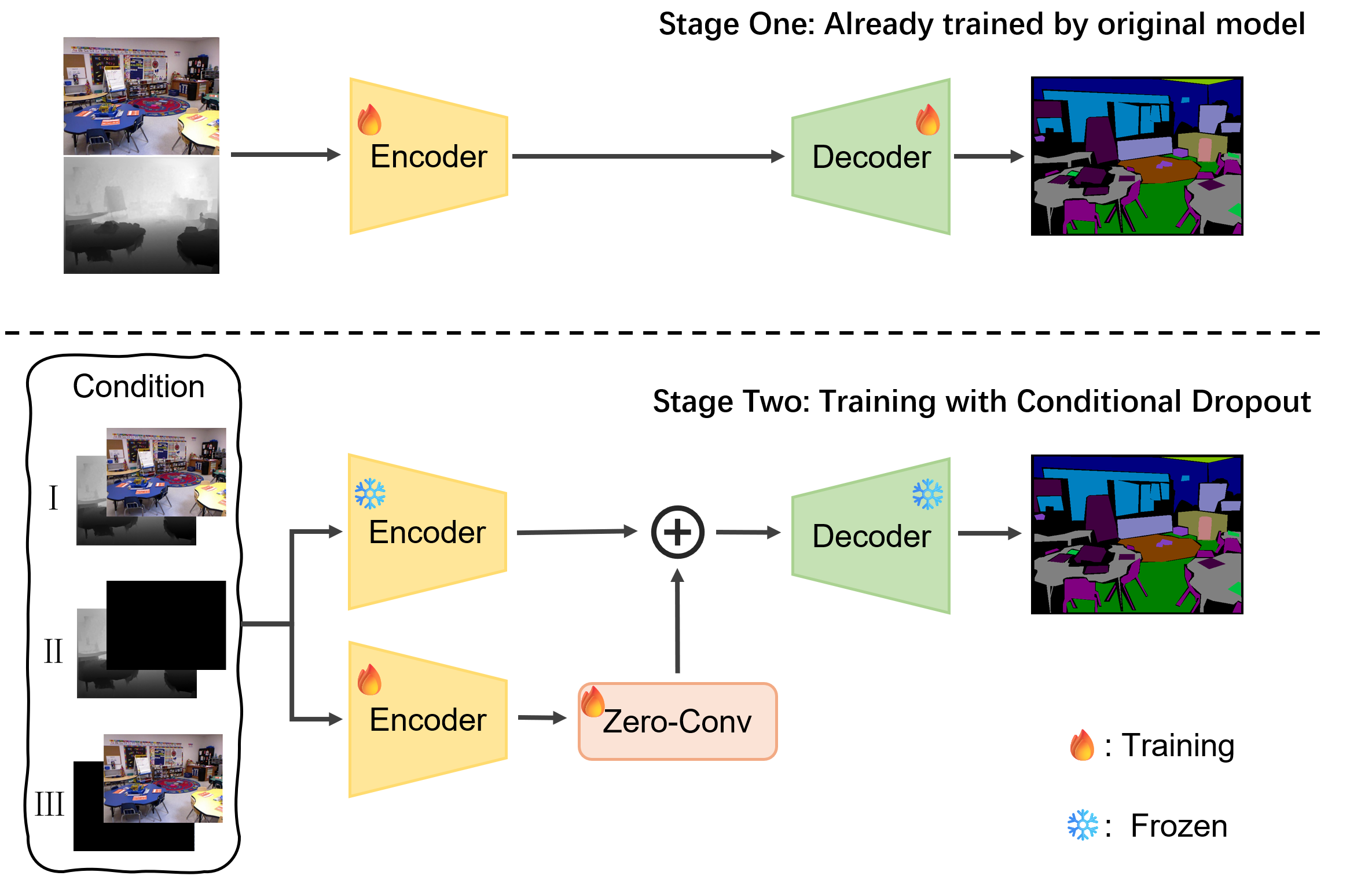}  % ensure the correct path
    \caption{
        Overview of the proposed conditional dropout training paradigm. 
\textbf{Stage~1 (existing model, no extra training):} any off-the-shelf RGB-D segmentation network that has been trained under the standard full-modality setting (Condition~I) can be directly adopted. We simply reuse its released checkpoint as initialization, without further modification or retraining. 
\textbf{Stage~2 (ours):} starting from this checkpoint, we perform a simple continued training with three input conditions (complete input, RGB-missing, depth-missing). During this stage, all parameters of the original network (encoders and decoder) are frozen, and a copy of each encoder is instantiated and updated. The copied-encoder outputs are injected via zero-initialized $1{\times}1$ convolutions and added to the frozen-encoder features before decoding. 
\emph{Only Stage~2 is trained in our pipeline}, which equips the model with robustness to missing modalities while preserving full-modal performance and incurring only minor additional computational cost.
    }
    \label{fig:framework}
\end{figure}

During continued training, $E_\text{pre}$ remains frozen to preserve previously learned full-modal representations, while $E_\text{aux}$ is optimized to adapt to partial-modal inputs. \emph{Since Stage~1 is reused without re-training, our method only performs Stage~2, substantially reducing compute and wall-clock cost compared with training the whole model from scratch.} An overview of the two-stage process is illustrated in Fig.~\ref{fig:framework}.

\subsection{Modality-Stochastic Input Strategy}

To simulate real-world variability in sensor availability, we introduce a stochastic input mechanism. Specifically, during each training iteration, the model receives one of three possible modality configurations with equal probability:

\begin{equation}
(x_{\text{RGB}}, x_{\text{D}}) \in \left\{
\begin{aligned}
    &(x_{\text{RGB}},\ x^{-}_{\text{D}}) \\
    &(x^{-}_{\text{RGB}},\ x_{\text{D}}) \\
    &(x_{\text{RGB}},\ x_{\text{D}})
\end{aligned}
\right\}, \quad P = \frac{1}{3}
\end{equation}

Here, $x^{-}_{\text{RGB}}$ and $x^{-}_{\text{D}}$ are symbolic placeholders indicating the absence of the RGB or depth modality, respectively. In practice, they are instantiated as zero tensors that match the shape of valid inputs.
% but are excluded from meaningful feature computation in the frozen encoder.

\subsection{Zero Convolution Feature Injection}

To allow the trainable encoder to influence the model without destabilizing the frozen backbone, we adopt a residual feature injection mechanism. The features extracted from $E_\text{aux}$ are first passed through a $1 \times 1$ convolution layer $\text{Conv}_0$ whose parameters are initialized to zero, similar to the approach in ControlNet~\cite{zhang2023adding}:
\begin{equation}
\hat{F} = \text{Conv}_0(E_\text{aux}(x_{\text{RGB}}, x_{\text{D}}))
\end{equation}
The resulting injected features are then added to the corresponding features from the frozen encoder:
\begin{equation}
F = E_\text{pre}(x_{\text{RGB}}, x_{\text{D}}) + \hat{F}
\end{equation}
This residual design ensures that the model output remains unchanged at the beginning of continued training. As training progresses, the injected features gradually encode useful information learned from partial-modal inputs, improving the model’s adaptability.

\subsection{Training Objective and Stability}

The final segmentation predictions are decoded from the fused feature $F$ using the original decoder $D_\text{pre}$, and the model is supervised using the same task-specific loss function (e.g., cross-entropy or panoptic quality). Importantly, our design maintains the inference architecture unchanged; only the training process is adapted. This makes our paradigm highly practical and compatible with a wide range of pre-existing RGB-D models.

By explicitly training under all possible modality configurations while leveraging frozen representations, the model gains robustness to missing modalities without sacrificing accuracy on complete inputs, \emph{while requiring only Stage~2 training in practice}.

\begin{table*}[ht]
\centering
\caption{Quantitative experiments of different RGB-D segmentation models in modality-complete and modality-missing conditions. ``w/o ConD'' indicates without Condition Dropout, while ``w/ ConD'' indicates with Condition Dropout. The evaluation metrics include $mIoU$ and $mAcc$, where higher values indicate better performance.}
\label{tab:main}
% \vspace{0.5em}
{\small
\begin{tabular}{c|cc|c|c|c|c|c}
\toprule
\multirow{2}{*}{\textbf{Datasets}} 
& \multicolumn{2}{c|}{\textbf{Conditions}}
& \multirow{2}{*}{\textbf{Metric}} 
& \textbf{DFormer-B}
& \textbf{DFormer-B}
& \textbf{Sigma-S}
& \textbf{Sigma-S}\\
&\textbf{RGB} &\textbf{Depth} & & \textbf{w/o ConD} & \textbf{w/ ConD} & \textbf{w/o ConD} & \textbf{w/ ConD}\\
\midrule
\multirow{10}{*}{\textbf{NYUv2}}
 & \multirow{2}{*}{$\mybullet$} & \multirow{2}{*}{$\mybullet$} & $mIoU$ & 55.6 & \textbf{55.7} \textcolor{red}{+0.1} & 56.6 & \textbf{57.2} \textcolor{red}{+0.6}\\
 & & & $mAcc$ & \textbf{69.4} & 68.6 \textcolor{blue}{-0.8} & 69.5 & \textbf{69.9} \textcolor{red}{+0.4}\\
 & \multirow{2}{*}{$\mycirc$} & \multirow{2}{*}{$\mybullet$} & $mIoU$ & 25.5 & \textbf{43.5} \textcolor{red}{+18.0} & 13.0 & \textbf{41.5} \textcolor{red}{+28.5}\\
 & & & $mAcc$ & 34.6 & \textbf{55.1} \textcolor{red}{+20.5} & 16.5 & \textbf{55.1} \textcolor{red}{+38.6}\\
 & \multirow{2}{*}{$\mybullet$} & \multirow{2}{*}{$\mycirc$} & $mIoU$ & 35.2 & \textbf{48.3} \textcolor{red}{+12.9} & 50.3 & \textbf{52.9} \textcolor{red}{+2.6}\\
 & & & $mAcc$ & 48.9 & \textbf{61.5} \textcolor{red}{+12.6} & 64.2 & \textbf{66.6} \textcolor{red}{+2.4}\\
\cmidrule{2-8}
 & \multicolumn{2}{c|}{\multirow{2}{*}{\textbf{Average Drop}}} & $mIoU$ & -25.3 & \textbf{-9.8} \textcolor{red}{+15.5} & -25.0 & \textbf{-10.0} \textcolor{red}{+15.0}\\
 & & & $mAcc$ & -27.7 & \textbf{-10.3} \textcolor{red}{+17.4} & -29.2 & \textbf{-9.1} \textcolor{red}{+20.1}\\
 & \multicolumn{2}{c|}{\multirow{2}{*}{\textbf{Average}}} & $mIoU$ & 38.8 & \textbf{49.2} \textcolor{red}{+10.4} & 40.0 & \textbf{50.5} \textcolor{red}{+10.5}\\
 & & & $mAcc$ & 46.4 & \textbf{61.7} \textcolor{red}{+15.3} & 50.1 & \textbf{63.9} \textcolor{red}{+13.8}\\
\midrule
\multirow{10}{*}{\textbf{SUN}} 
 & \multirow{2}{*}{$\mybullet$} & \multirow{2}{*}{$\mybullet$} & $mIoU$ & 51.2 & \textbf{51.4} \textcolor{red}{+0.2} & 51.9 & \textbf{52.9} \textcolor{red}{+1.0}\\
 & & & $mAcc$ & \textbf{63.0} & 62.8 \textcolor{blue}{-0.2} & 64.0 & \textbf{64.6} \textcolor{red}{+0.6}\\
 & \multirow{2}{*}{$\mycirc$} & \multirow{2}{*}{$\mybullet$} & $mIoU$ & 13.6 & \textbf{38.1} \textcolor{red}{+24.5} & 11.6 & \textbf{37.8} \textcolor{red}{+26.2}\\
 & & & $mAcc$ & 17.6 & \textbf{47.1} \textcolor{red}{+29.5} & 14.5 & \textbf{47.8} \textcolor{red}{+33.3}\\
 & \multirow{2}{*}{$\mybullet$} & \multirow{2}{*}{$\mycirc$} & $mIoU$ & 41.7 & \textbf{45.3} \textcolor{red}{+3.6} & 46.0 & \textbf{50.1} \textcolor{red}{+4.1}\\
 & & & $mAcc$ & 51.9 & \textbf{55.8} \textcolor{red}{+3.9} & 57.4 & \textbf{62.7} \textcolor{red}{+5.3}\\
\cmidrule{2-8}
 & \multicolumn{2}{c|}{\multirow{2}{*}{\textbf{Average Drop}}} & $mIoU$ & -23.6 & \textbf{-9.7} \textcolor{red}{+13.9} & -23.1 & \textbf{-9.0} \textcolor{red}{+14.1}\\
 & & & $mAcc$ & -28.3 & \textbf{-11.4} \textcolor{red}{+16.9} & -28.1 & \textbf{-9.4} \textcolor{red}{+18.7}\\
 & \multicolumn{2}{c|}{\multirow{2}{*}{\textbf{Average}}} & $mIoU$ & 35.5 & \textbf{44.9} \textcolor{red}{+9.4} & 36.5 & \textbf{46.9} \textcolor{red}{+10.4}\\
 & & & $mAcc$ & 44.2 & \textbf{55.2} \textcolor{red}{+11.0} & 45.3 & \textbf{58.4} \textcolor{red}{+13.1}\\
\bottomrule
\end{tabular}
}
\end{table*}

%%%%%%%%%%%%%%%%%%%%%%%%%%%%%%%%%%%%%%%%%%%%%%%%%%%%%%%%%%%%%%%%%%%%%%%%%%%%%%%%%%%%%%%%%%%
\section{Experiments}

\subsection{Datasets and Evaluation Metrics}
\label{sec:datasets_metrics}
% We evaluate our method on two widely used RGB-D semantic segmentation benchmarks: NYU-Depth V2 and SUN RGB-D. NYU-Depth V2 consists of 1,449 RGB-D images, including 795 for training and 654 for testing, covering 40 object categories. SUN RGB-D contains 10,335 images, with 5,285 for training and 5,050 for testing, spanning 37 categories.
% For evaluation, we report the mean Intersection over Union (mIoU) and mean Accuracy (mAcc), which are commonly adopted to assess segmentation performance. Higher values indicate better results.
We evaluate our method on two standard RGB-D semantic segmentation benchmarks: NYU-Depth V2~\cite{silberman2012indoor} and SUN RGB-D~\cite{song2015sun}. NYU-Depth V2 contains 1,449 images (795 for training, 654 for testing) with 40 categories, while SUN RGB-D provides 10,335 images (5,285 training, 5,050 testing) across 37 categories. Performance is measured by mean Intersection over Union (mIoU) and mean Accuracy (mAcc), where higher values indicate better segmentation quality.

\subsection{Implementation Details}
\label{sec:implementation}

All models are implemented in PyTorch and trained on two NVIDIA RTX 3090 GPUs with input resolution of $480 \times 640$. To ensure fair comparison, we follow the original papers for all hyperparameter settings, including optimizer, learning rate, and training schedules. Standard data augmentations such as random flipping and color jittering are applied during training.

\subsection{Comparison with State-of-the-Art}
\label{sec:sota_comparison}

% We compare our method with existing state-of-the-art RGB-D segmentation models under both modality-complete and modality-missing conditions. As shown in Table~\ref{tab:main}, our method consistently improves performance in missing-modality scenarios without compromising the results when both modalities are present. Notably, our model achieves substantial gains in $mIoU$ and $mAcc$ under degraded conditions, especially on the NYU-Depth V2 dataset.

% Surprisingly, even under the modality-complete setting, our method maintains competitive performance and in some cases achieves slight improvements. This suggests that the proposed training strategy not only enhances robustness to missing modalities but also has a positive regularization effect, potentially improving the model's representation ability in full-modality scenarios.

\begin{figure*}[t]
    \centering
    \includegraphics[width=\textwidth]{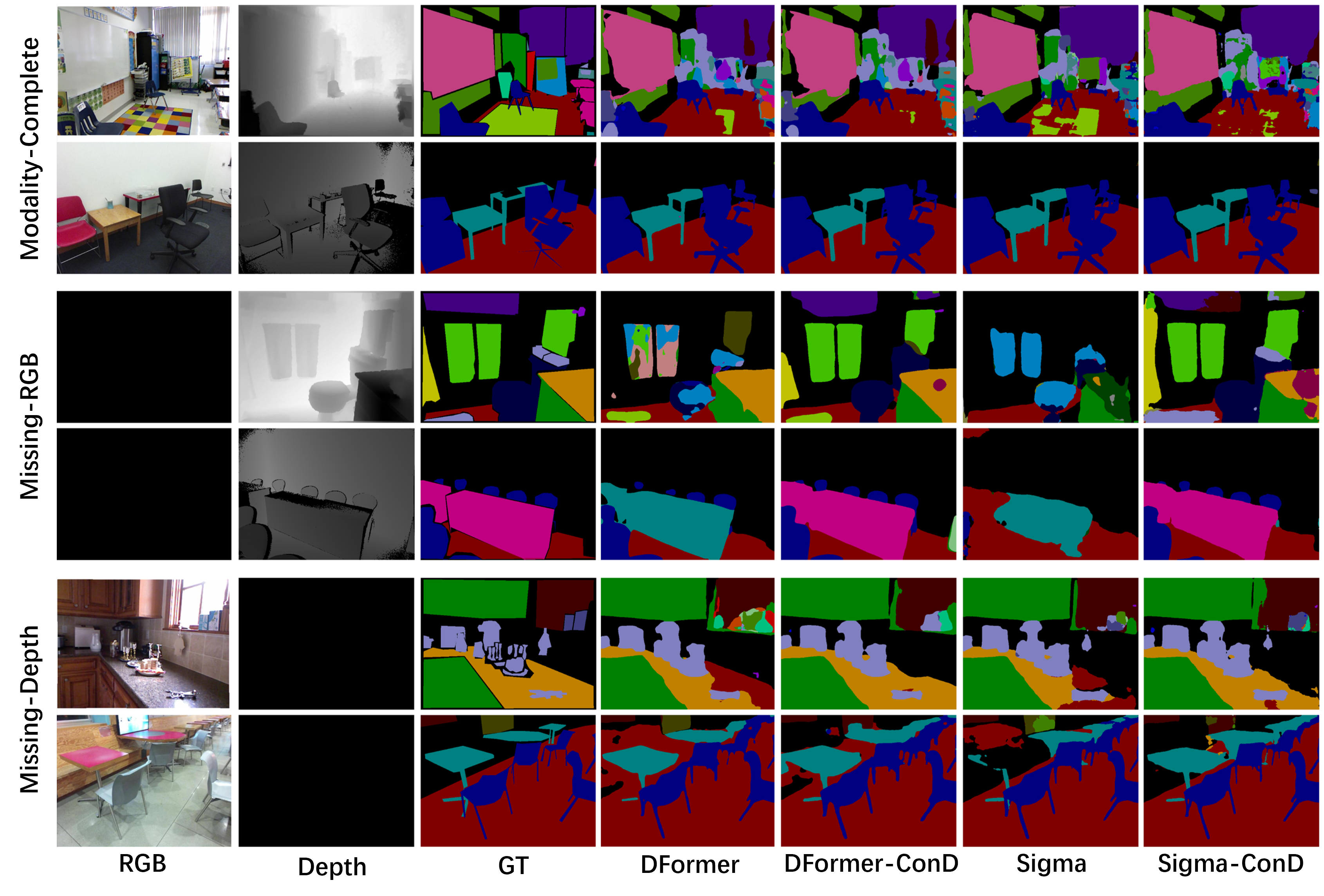}
    \caption{Qualitative comparison of segmentation results under different modality conditions. We present results from two representative RGB-D segmentation models, DFormer and Sigma, both with and without the proposed Condition Dropout. It can be observed that after introducing ConD, both models achieve improved segmentation quality not only under the complete-modality setting but also under missing-modality inputs (RGB-missing or Depth-missing).}
    \label{fig:Qualitative}
\end{figure*}
To further validate the effectiveness of our training strategy, we apply Condition Dropout to representative state-of-the-art RGB-D segmentation models, including DFormer-B~\cite{yin2023dformer} and Sigma-S~\cite{wan2025sigma}, and conduct evaluations on both NYUv2 and SUN benchmarks. As presented in Table~\ref{tab:main}, incorporating ConD consistently leads to significant performance improvements under modality-missing scenarios. In particular, the average performance drop caused by single-modality input is dramatically alleviated: for example, on NYUv2, the $mIoU$ drop is reduced from $-25.3$ to $-9.8$ for DFormer-B, and from $-25.0$ to $-10.0$ for Sigma-S. A similar trend is observed on SUN, where the drop in $mAcc$ shrinks from more than $-28.3$ to below $-11.4$. In other words, ConD turns these models from being highly fragile under missing modalities into much more robust systems that still produce reasonable predictions even when one modality is unavailable.

\begin{figure}[ht]
    \centering
    \includegraphics[width=\linewidth]{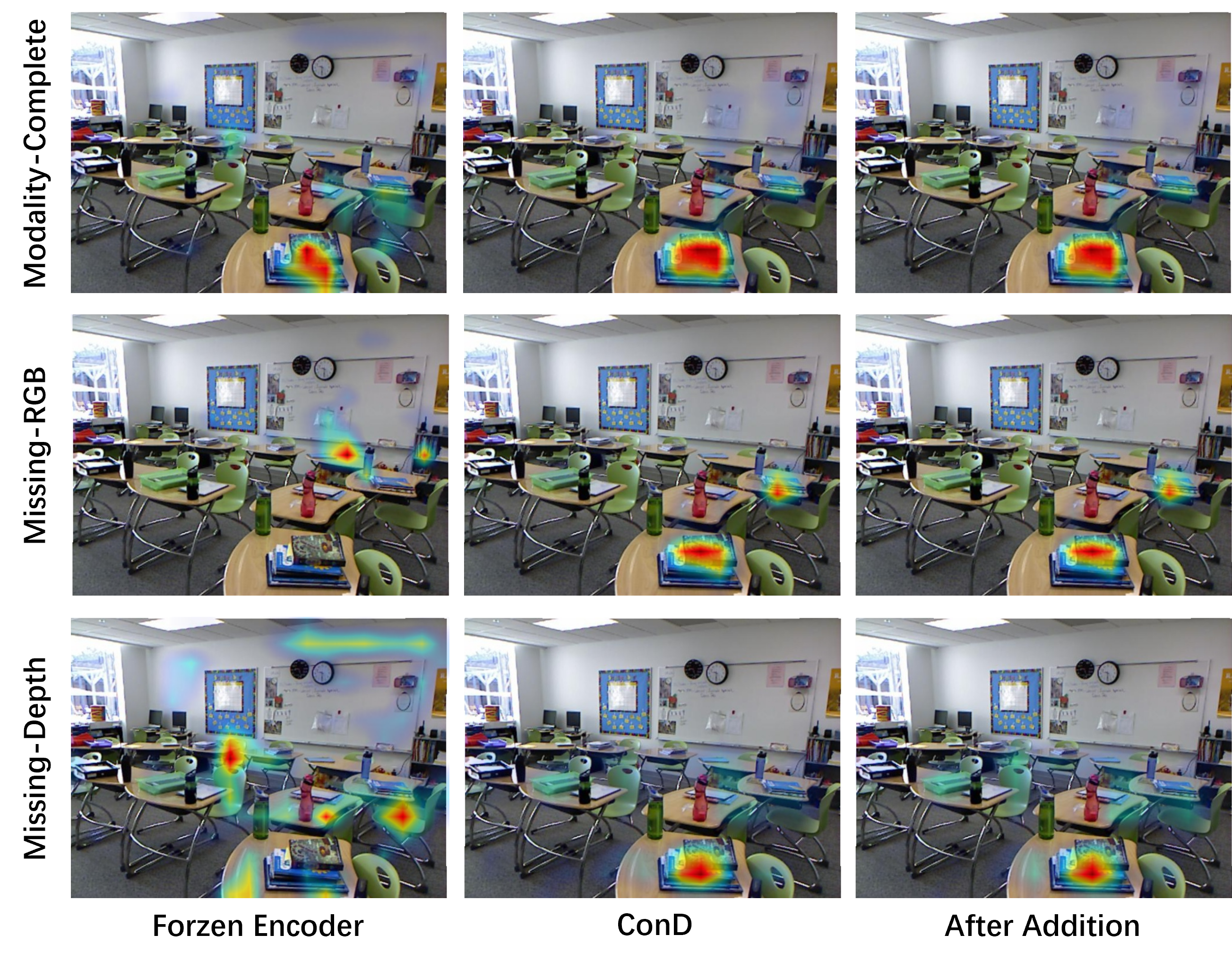}
    \caption{Visualization of feature maps from the original encoder and the auxiliary encoder under different modality conditions. The auxiliary encoder consistently provides complementary semantic cues, particularly in regions where the original encoder exhibits weak or ambiguous responses. These enriched features contribute to improved robustness under missing modalities and enhanced representation quality under complete-modality inputs.}
    \label{fig:feature_vis}
\end{figure}

It is also worth noting that ConD not only enhances robustness under missing modalities but also provides consistent gains in modality-complete conditions. For instance, on NYUv2 and SUN, both DFormer-B and Sigma-S exhibit improvements of up to +1.0 in $mIoU$ and +0.6 in $mAcc$ when all modalities are present. Moreover, our feature visualizations in Fig.~\ref{fig:feature_vis} show that the encoder outputs become more discriminative with ConD, under both complete and missing inputs, confirming that the model learns to better exploit the available modality information. This indicates that although the primary objective of ConD is to mitigate performance degradation in incomplete-modality settings, it simultaneously strengthens feature extraction for each modality, thereby yielding additional improvements in the full-modality scenario. Overall, this comparison confirms that our method can be seamlessly integrated into advanced segmentation frameworks and substantially improve both their resilience and their representational capacity.

\begin{table*}[ht]
\centering
\caption{Ablation study of four configurations: (A) baseline, (B) with modality dropout (MD), (C) with MD and encoder duplication (Copy) but without freezing, and (D) the full method with MD, Copy, and freezing (Freeze). Results show that only the full combination preserves full-modality accuracy while substantially improving robustness under missing modalities.}
\begin{tabular}{cccc|cc|cc|cc|cc}
\toprule
\multirow{2}{*}{}& \multirow{2}{*}{MD}& \multirow{2}{*}{Copy} & \multirow{2}{*}{Freeze} & \multicolumn{2}{c|}{Modality Complete} & \multicolumn{2}{c|}{Missing RGB} & \multicolumn{2}{c|}{Missing Depth} & \multicolumn{2}{c}{Average} \\
 && & & $mIoU$ & $mAcc$ & $mIoU$ & $mAcc$& $mIoU$ & $mAcc$ & $mIoU$ & $mAcc$ \\
\midrule
(A) & & &   &51.2 &63.0 &13.6 &17.6 &41.7 &51.9 &35.5 &44.2  \\
(B) &$\checkmark$  && &47.5 &57.3 &25.7 &29.0 &43.5 &52.7 &38.9 &46.3  \\
(C) &$\checkmark$  &$\checkmark$& &48.8 &58.9&27.9 &34.8 &44.1 &54.0 &40.3 &48.6 \\

\midrule
(D) &$\checkmark$  &$\checkmark$ &$\checkmark$  &\textbf{51.4} &\textbf{62.8} &\textbf{38.1} &\textbf{47.1} &\textbf{45.3} &\textbf{55.8} &\textbf{44.9} &\textbf{55.2}  \\
\bottomrule
\end{tabular}
\label{table_Ablation}
\end{table*}

\subsection{Qualitative Analysis}

In addition to quantitative evaluations, we further provide qualitative comparisons to illustrate the benefits of our proposed Condition Dropout. Representative results are shown in Fig.~\ref{fig:Qualitative}, covering three input scenarios: complete RGB-D inputs, RGB-missing, and depth-missing. 

From the visualization, we observe that baseline models such as DFormer and Sigma often fail to generate coherent segmentation masks under missing-modality conditions, leading to incomplete object boundaries and misclassified regions. After incorporating ConD, both models produce substantially more consistent and accurate predictions across all settings. In particular, ConD helps the networks better preserve semantic structures when only a single modality is available, while also maintaining high-quality predictions under the complete-modality case. 

These results confirm that the proposed training paradigm not only enhances robustness to modality incompleteness but also improves the overall reliability of RGB-D segmentation models in diverse real-world scenarios.

\subsection{Feature Visualization and Analysis}

To gain a clearer understanding of how the proposed Condition Dropout strategy affects feature representations, we visualize and compare the feature maps produced by the original encoder and the auxiliary encoder under different modality conditions. The visual results reveal that the auxiliary encoder progressively learns complementary representations during Stage~2 training, enabling it to effectively supply semantic information that becomes missing or unreliable when one input modality is absent.

Specifically, under complete-modality inputs, the original encoder produces stable and well-structured activations. When one modality is removed, however, the remaining modality still contains sufficient semantic cues for reasonable segmentation, as can be observed from the input images. Yet the baseline encoder’s responses in this case deteriorate noticeably, with weakened activations, fragmented structures, and missing object regions, revealing its lack of robustness to modality missing. After introducing ConD, the encoder responses under the same missing-modality conditions are much better preserved: the activations remain aligned with object shapes and boundaries and cover most of the semantic regions that were lost in the baseline. Through the zero-initialized residual injection mechanism, these improved responses are gradually fused into the frozen encoder’s representation, yielding more complete and robust feature maps even when a modality is absent.

More importantly, even under complete-modality inputs, the auxiliary encoder still produces features that are non-redundant yet semantically aligned with those of the original encoder. This indicates that Stage~2 training not only enables the model to remain reliable under modality-missing scenarios, but also encourages the two encoders to form a more expressive joint feature space. This phenomenon aligns well with our empirical findings: Condition Dropout substantially mitigates performance drops under missing modalities, while also yielding consistent gains under full-modality inputs (Table~\ref{tab:main}). The improvements are further validated through ablation studies (Table~\ref{table_Ablation}), confirming that the auxiliary encoder provides genuinely useful semantic information rather than merely benefiting from an increased parameter count.

Overall, the feature visualizations validate the core design principle of our method: the duplicated encoder, without altering the frozen backbone’s behavior, learns complementary and structurally meaningful representations. Together, the two encoders significantly enhance the robustness and expressiveness of the segmentation model across different input conditions.

\subsection{Ablation Study}
\label{sec:ablation}

To evaluate the effectiveness of our training paradigm, we conduct an ablation study with four configurations: (a) the baseline trained only on full-modality inputs; (b) the baseline with stochastic modality dropout but without encoder duplication; (c) the model with encoder duplication and zero-convolution injection but without freezing the original encoder; and (d) our full method—Condition Dropout.

As shown in Table~\ref{table_Ablation}, comparing (a) and (d) highlights the overall effectiveness of ConD: the baseline collapses under missing modalities, while ConD achieves robustness without sacrificing full-modality performance. The gap between (b) and (d) shows that dropout alone hurts complete inputs, and the difference between (c) and (d) confirms the necessity of freezing the original encoder. These results demonstrate that dropout, duplication, and freezing are complementary, and only their combination yields the best balance between robustness and accuracy.

\subsection{Limitations}
\label{sec:limitations}
Although ConD is simple and effective, it still has limitations. Introducing copied encoders inevitably increases the number of parameters and memory consumption, and these extra encoders are also used at inference time, leading to higher latency than the original backbone. This overhead may be problematic for highly resource-constrained or real-time applications. Future work will explore more parameter- and computation-efficient variants of ConD, such as lightweight adapters, partial encoder sharing, or distillation into a compact single-encoder model.

%%%%%%%%%%%%%%%%%%%%%%%%%%%%%%%%%%%%%%%%%%%%%%%%%%%%%%%%%%%%%%%%%%%%%%%%%%%%%%%%%%%%%%%%%%%
\section{Conclusion}
This paper addresses the pervasive challenge of modality missing in RGB-D semantic segmentation by proposing an innovative continuous training strategy: Conditional Dropout. This method enhances the model's ability to handle incomplete modal inputs through a continued training strategy on a pre-trained model. Experimental results demonstrate a significant improvement in robustness under modality-missing scenarios, while maintaining or even slightly surpassing the original performance under complete modal inputs. The ConD strategy strengthens the overall representational capacity of the model by enhancing feature extraction and can be seamlessly integrated into existing advanced segmentation frameworks.

Despite the excellent performance of the ConD strategy in RGB-D semantic segmentation, its universality in other multimodal tasks still requires further validation. Furthermore, its effectiveness is contingent on the availability of high-quality pre-trained models. Future research could explore more generalized mechanisms for handling modality missing and optimize the ConD strategy for resource-constrained or data-scarce environments.

\bibliographystyle{IEEEbib}
\bibliography{icme2026references}

\end{document}